\documentclass[sn-mathphys]{sn-jnl}

\jyear{2021}%

\theoremstyle{thmstyleone}%
\theoremstyle{thmstyletwo}%

\theoremstyle{thmstylethree}%

\bibliography{ref}
\usepackage{multirow}
\usepackage {mathtools}
\begin{document}

\title[Monocular Depth Estimation with Sharp Boundary]{Monocular Depth Estimation with Sharp Boundary}


\author[1,2]{\fnm{Xin} \sur{Yang}}\email{yx942411526@gmail.com}

\author[2,3]{\fnm{Qingling} \sur{Chang}}\email{qlchangcas@gmail.com}

\author[1,2]{\fnm{Xinlin} \sur{Liu}}\email{jmxlliu@163.com}
\author*[1,2,3]{\fnm{Yan} \sur{Cui}}\email{cuiyan@wyu.edu.cn}

\affil[1]{\orgdiv{Faculty of Intelligent Manufacturing}, \orgname{Wuyi University}, \city{Jiangmen}, \postcode{529000},  \country{China}}

\affil[2]{\orgdiv{China-Germany (Jiangmen) Artificial Intelligence Institute},   \city{Jiangmen}, \postcode{529000},  \country{China}}

\affil[3]{\orgdiv{Zhuhai 4Dage Network Technology}, \city{Zhuhai}, \postcode{519000},  \country{China}}


\abstract{Monocular depth estimation is the base task in computer vision. It has a tremendous development in the decade with the development of deep learning. But the boundary blur of the depth map is still a serious problem. Research finds the boundary blur problem is mainly caused by two factors, first, the low-level features containing boundary and structure information may loss in deeper networks during the convolution process., second, the model ignores the errors introduced by the boundary area due to the few portions of the boundary in the whole areas during the backpropagation. In order to mitigate the boundary blur problem, we focus on the above two impact factors. Firstly, we design a scene understanding module to learn the global information with low- and high-level features, and then to transform the global information to different scales with our proposed scale transform module according to the different phases in decoder. Secondly, we propose a boundary-aware depth loss function to pay attention to the effects of the boundary’s depth value. The extensive experiments show that our method can predict the depth maps with clearer boundaries, and the performance of the depth accuracy base on NYU-depth v2 and SUN RGB-D is competitive.}

\keywords{Monocular depth estimation, object boundary, blurry boundary, Scene information, Feature fusion, Scale transform, Boundary aware}



\maketitle

\begin{figure}[h]%
\centering
\includegraphics[width=0.9\textwidth ]{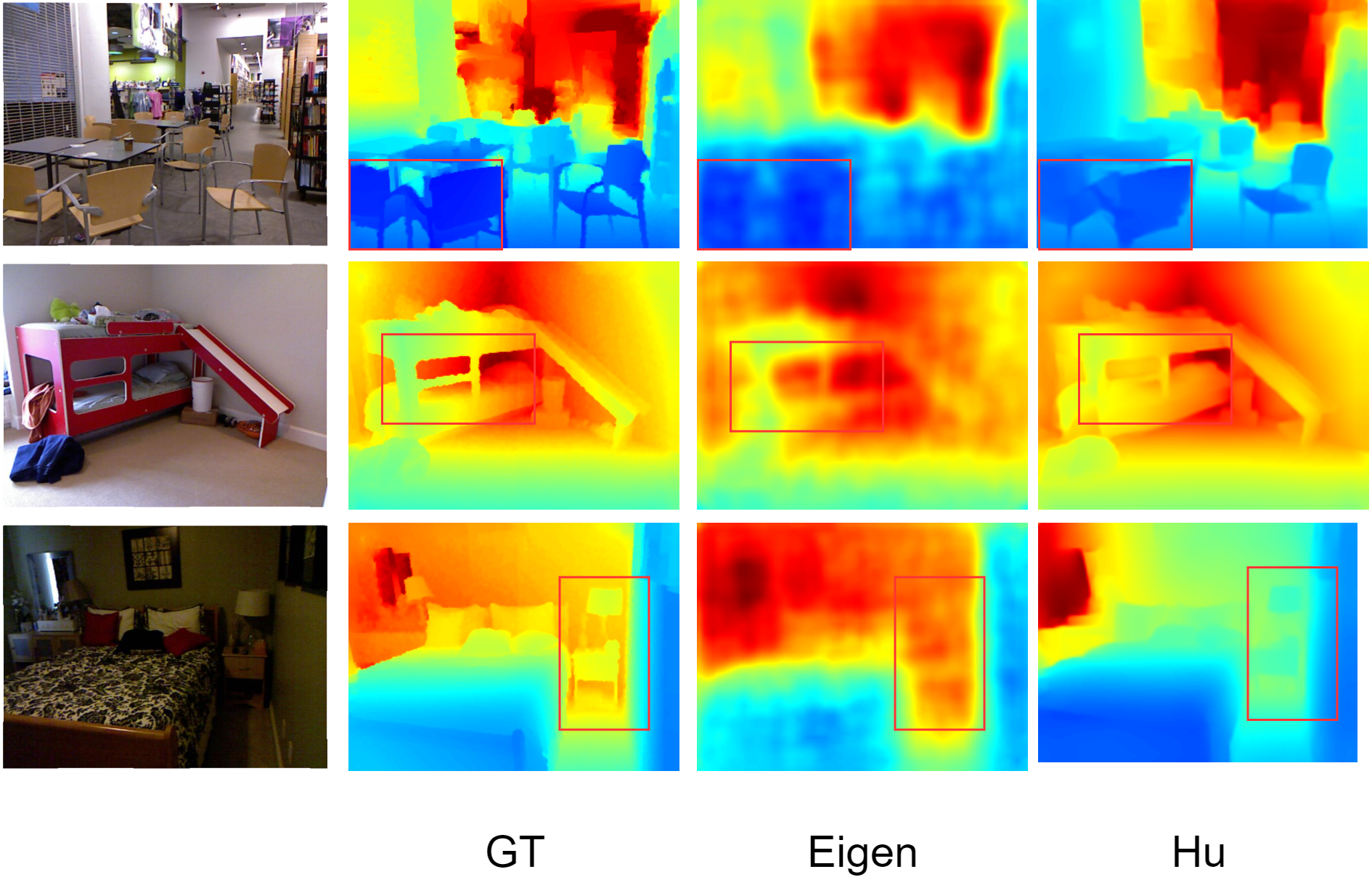}
\caption{Predicted depth map from other methods. From left to right are the input image, Ground truth, and the predicted from Eigen et al. [3] and Hu et al. [10]}\label{fig1}
\end{figure}

\begin{figure}%
\centering
\includegraphics[width=0.9\textwidth ]{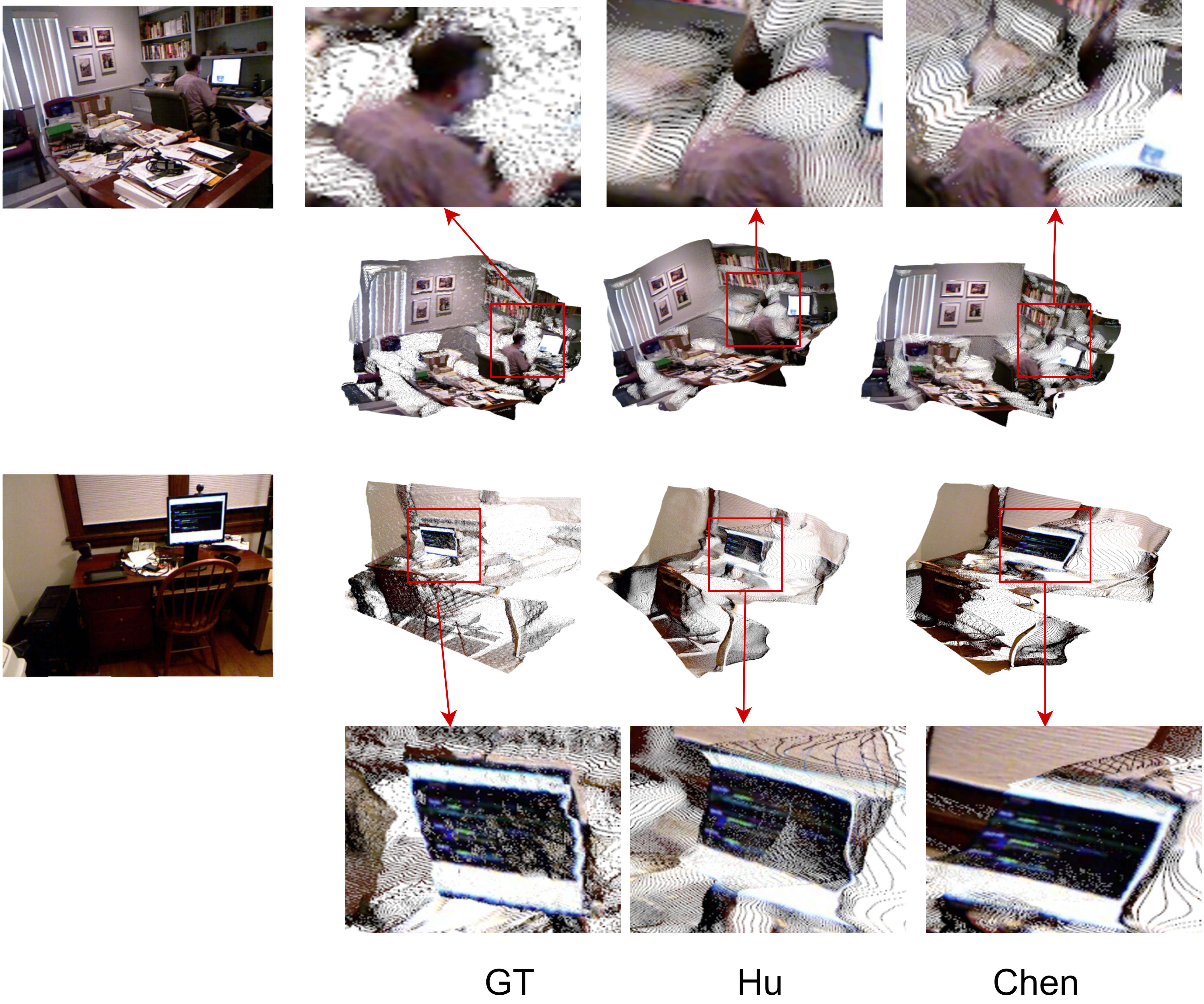}
\caption{The flying pixels phenomenon in the projected cloud from depth maps. From left to right are input RGB images, ground-truth depth maps, results of Hu et al. [10], and Chen et al. [13] respectively.}\label{fig2}
\end{figure}
\section{Introduction}\label{sec1}
Monocular depth estimation is a base vision task in computer vision. It is widely used in autonomous driving, height measurement, SLAM (Simultaneous Localization And Mapping), and AR (Augmented Reality), etc. Monocular depth estimation is denser and low-cost than traditional cost sensor to obtain depth directly and has advantages of low price, rich information acquisition content, and small size compared with the binocular image which is limited by the baseline length resulting in a poor match between the volume of the equipment and the vehicle platform. So, estimating depth information based on monocular cameras is of great significance and has become one of the research hotspots in the field of computer vision. Monocular depth estimation refers to transforming a 2D RGB image to a 2.5D depth map, which relies on the “Shape From X” methods to obtain the scene depth information from an RGB image. Monocular depth estimation is an ill-posed problem for a single image that lacks geometrical information to inference the depth from the image. Compared with the traditional methods [1] [2] etc using the cues designed artificially, learning-based monocular depth estimation methods develop more quickly and widespread profit from the Convolutional Neural Networks (CNNs) to extract features from images instead of the artificial cues, and then build the mapping between features and depth. Eigen et al. [3] proposed the first monocular depth estimation method base on deep learning, which showed a surprising performance than pre-works [1] [2]. Then, a lot of excellent works based on deep learning were proposed, such as [4] [5] [6] [7] [8] [9] [10] [11]. However, monocular depth estimation methods still suffered from the boundary blur challenge, especially in indoor scenes which have complex scene structures and many objects. Figure 1 shows the estimation depth results of some existing works in indoor scenes.
From the red box in Figure 1, we can see  Hu et al. [10] has a significant improvement in edge definition compared with Eigen et al.[3], but there is still a very obvious boundary blur phenomenon, especially in the complex object structure. The blurry boundary not only introduces predicting errors to the depth estimation but also causes the " flying pixels " [12]. Some cases about “flying pixels” in the predicted depth maps can be seen in Figure 2.  Look the Figure 2, the first group shows that the point clouds from Hu et al. [10],  and Chen et al. [13] have serious “flying pixels” in the human head part of the images. In the second group, the screens have wrapped. The projected point clouds from depth maps are discontinuous, especially in the object boundary. The depth values of pixels from boundary and non-boundary are different when projected to point clouds. Pixels with different depth values will be projected to different planes, leading to the boundary and non-boundary of one object being separated. Studies find that the boundary blur problem is mainly caused by two factors in the CNN frameworks. First, the loss of low-level features in the encode phase. But the low-level feature includes scene structure and object information, the deeper the network, the worse of the loss. But the deeper network is needed to improve the spatial expression ability and receptive field of the features due to the deeper network can extract the high-level and abstract information of images. Second, the "boundary smoothing" during model training (Boundary smoothing: the loss caused by boundary area is ignored during model training due to the boundary area occupies a little proportion, although the gradient of boundary is larger, it causes little loss in training. Processing the boundary as the non-boundary with a small gradient leads the boundary not sharp or clear enough, like the methods [10][13] in Figure 1. To solve the boundary blur problem, we propose two solutions for the two impact factors respectively: firstly, to mitigate the low -level information loss, we propose a Scene Understanding model (SU) and a Scale Transform (ST) module. Secondly, to solve the problem caused by “boundary
smoothing”, we pay attention to the loss introduced by the boundary and design a novel depth loss function Boundary Aware Depth loss (BAD).\\
\textbf{SU and ST}. Low-level features are easily lost in deep network, we can fuse the low-level features to the high-level features, the Fuse Feature Pyramid (FFP) is an usual method to aggregate different scales, such as Chen et al. [13], Yang et al. [14], but these models have a lot of params due to sampling the features too many times during building the FFP. To reduce the params, SU and ST are designed in this work. SU is used to aggregate all scales feature extracted from all encode phases with channel compression operation on every scale and learning the scene information needed in the decoder phases, and it the outputs the global scene information-rich of scene structure and boundary and abstract information Then, ST transforms the global scene information to different scales according to the corresponding phases in the encoder, and connects each corresponding scales in encoder and decoder with skip connection. What’s more, we also design an adaptive learning mechanism in ST to learn the needed scale according to the corresponding scale in the encoder.\\
\textbf{BAD.} BAD introduce the boundary weights to the depth loss function, the weight composed of multi-items to ensure it is useable in most cases. The boundary area will be set a bigger weight in the loss function to alleviate the smoothing phenomenon caused by too little proportion of the border area.
   Our contributions can be summarized as follows:\\
1.	We propose a Scene Understanding module (SU) to aggregate the multi-scale features of the encoder and learn the global scene information, furthermore, we  design a Scale Transform module (ST) to transform global information to different scales, which transform feature and resolution adaptively to suit each decode phase with learning mechanism\\
2.	We propose a novel Boundary Aware Depth loss (BAD). BAD introduces a boundary-aware weight to the depth loss and leads the model to aware of their pixels which have high edge gradients. \\
3.	Extensive experimental results show that our model can predict accurate depth value, clearer boundary, which can effectively alleviate " flying pixels " and has competitive depth accuracy in the NYU-Depth V2 and SUN RGB-D dataset.

\section{Related work}\label{sec2}

Monocular depth estimation is an important task in computer vision. The beginning of the learning-based monocular depth estimation is Eigen et al. [3], which predicted the depth from a single RGB image. Eigen et al. [3] shows a great performance over  the previous works [1] [2]. Base on this paper, Eigen et al. [4] proposed a universal multi-task framework to predict depth, surface normal, and segments, from a single image. After that, the monocular depth estimation has made great progress. And then, some researchers proposed to fuse the CRF(Conditional Random Field) and deep learning [15] [16] [17] [18]. The combining CRF and CNNs makes up for the problems of CNNs and improves the accuracy of the depth estimation models. In addition, [9] [11] [19] proposed to use classification to deal with the monocular depth estimation. By dividing the depth interval of the image, by solving the pixel interval corresponding to each pixel and using the depth value corresponding to the depth interval to express the depth of each pixel. These works focus on accuracy, but ignore the structure information in depth map, which will impact the effects of reconstruction or obstacle detection with point clouds projected from depth map without or with less structure information,especially in the complex scene such as the indoor scenes, which have complicated structure and a mass of objects. Clear object boundary not only improves accuracy, but also keeps the object good shape. But most of the existing CNN model may loss the low-level feature or structure information,leading to the boundary problem. To deal with the blurry boundary, Hu et al. [10] proposed a fusion model to fuse multi-scale features and proposed a compound loss function to make the boundary clearer. Base on this excellent work, Chen et al. [13] proposed a Fused Feature Pyramid(FFP) and a residual pyramid to predict depth map. Base on Chen et al. [13],  Yang et al. [14] built a FFP  and used a ASFF(Adaptively Spatial Feature Fusion) structure [20] to fuse the different scale depth maps to keep the structure information. Although Chen et al. [13] and Yang et al. [14] show a great performance, but the model is too big and the params are too many for the FFP. So, we propose a SU model to fuse the multi-scale feature with channel compression and propose a ST module to transform the global scene information to according to scale with current phase of decoder through learning mechanism. and during which, we only need to implement multi-scale feature fusion at SU once, and then obtain fusion features of different resolutions through ST adaptively. Furthermore, to predict the depth include clearer object boundary, we propose a novel depth loss BAD according to the different gradients of the region, with the corresponding depth loss weight to enforce the network to punish the depth error in boundary field. Additional punishing will ensure more accuracy depth predicting in the boundary field with more shape object boundary.
\section{Proposed method}\label{sec3}
In this section, we first introduce the overall framework, and then we describe the Scene Understanding module (SU), the Scale Transform module (ST), and Boundary Aware Depth loss (BAD) in detail successively.

\begin{figure}%
\centering
\includegraphics[width=0.9\textwidth ]{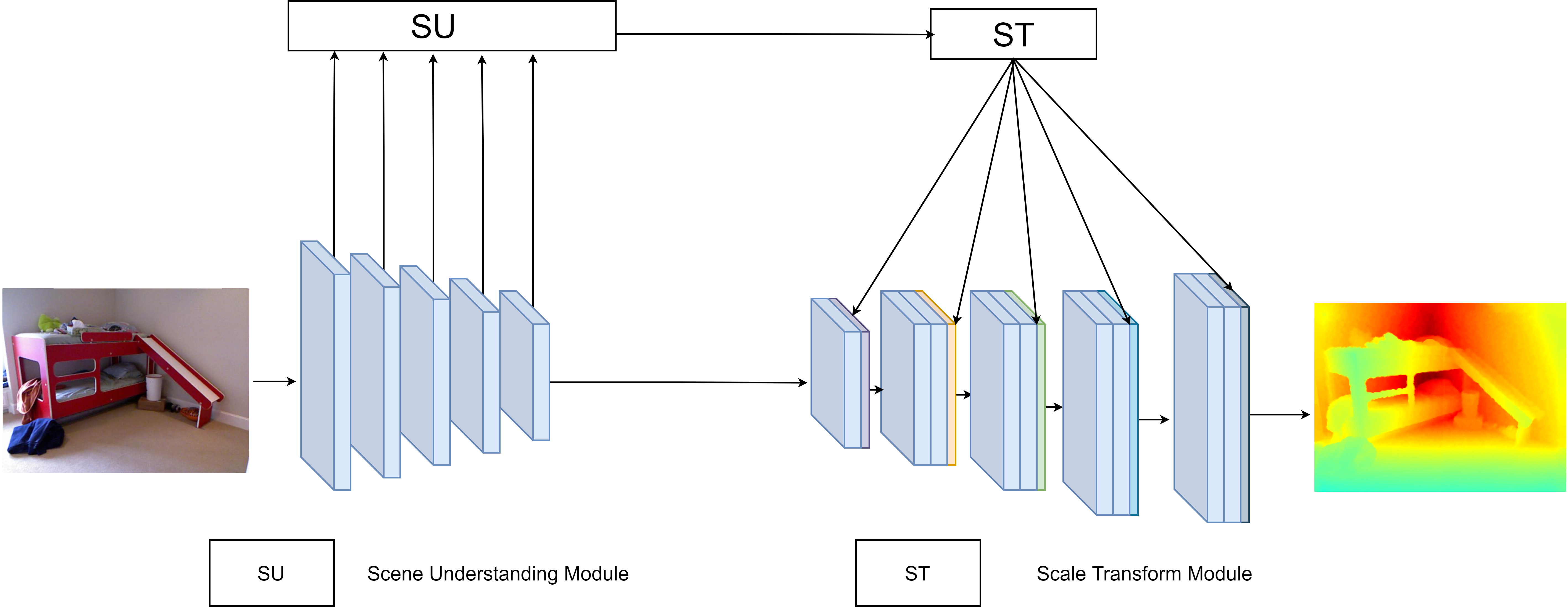}
\caption{The architecture of our framework. SU is meaning the Scene understanding module and the ST is the Scale Transform module.}\label{fig3}
\end{figure}
\subsection{Overall framework}\label{subsec3}
Blurry boundary not only introduces error in-depth maps but also causes “flying pixel” in point clouds, to alleviate the boundary blur problem, we design the framework shown in Figure 3, which uses the encoder-decoder architecture and select the SeNet154 [21] as the backbone. The framework mainly focuses on the two main factors causing the blurry boundary in-depth maps: 1) the low-level feature loss; 2) the boundary smooth strategy. In the framework, besides the base encoder-decoder architecture, we design SU model in the encoder stage to aggregate all extracted features with channel compression and also learn the global scene information, and ST model in the decoder stage to transform the global information to different scales. Furthermore, to deal with the problem introduced by boundary smooth, we propose a novel loss function Boundary Aware Depth loss (BAD) to enforce the model to notice the object boundary in the training process but not ignore it directly.
\begin{figure}%
\centering
\includegraphics[width=0.9\textwidth ]{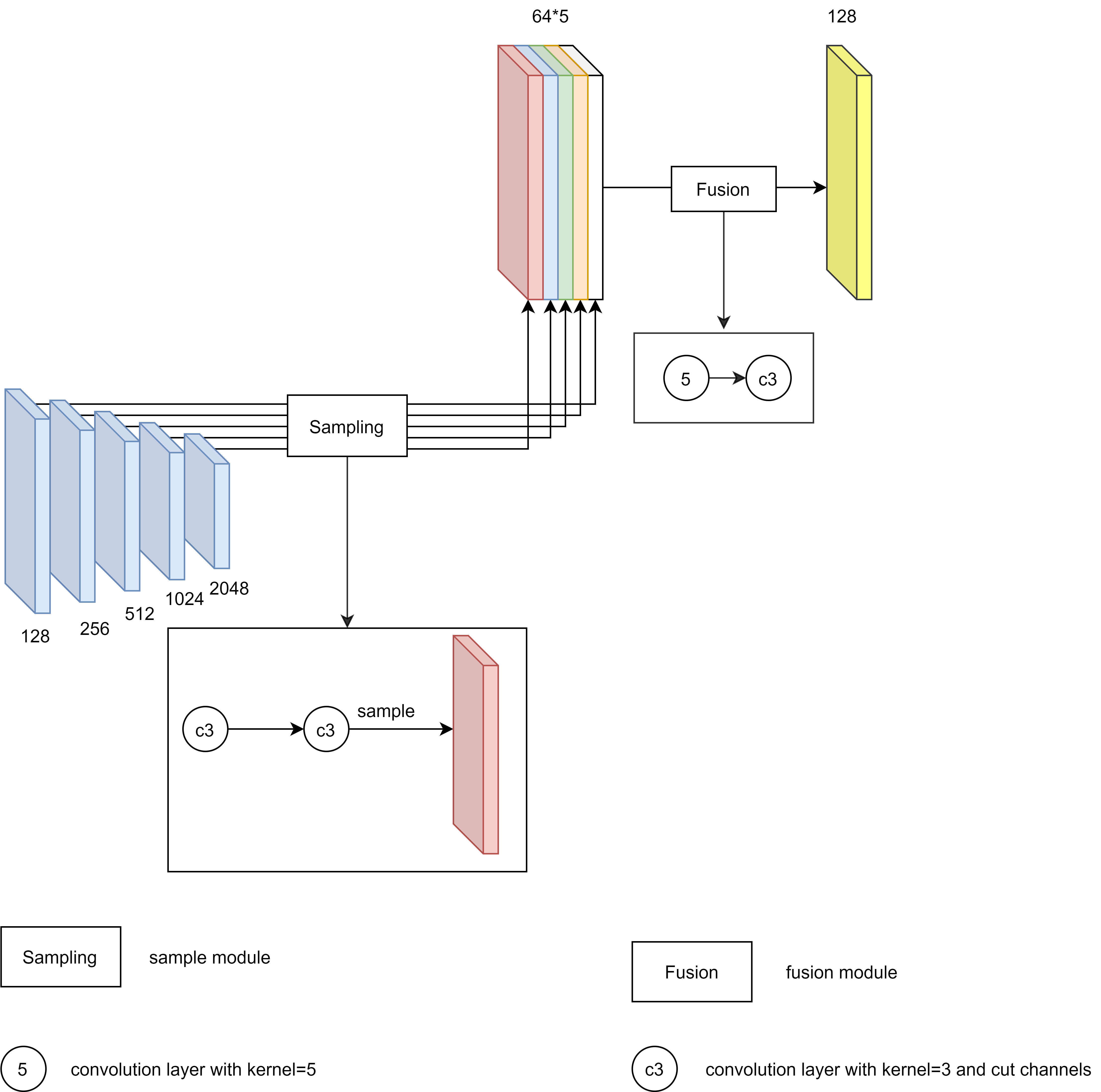}
\caption{The architecture of the Scene Understanding model. The sample module uses to sampling each scale feature to the same scale and the fusion module is used to adaptive fuse multi-scale features for learning the whole scene feature}\label{fig4}
\end{figure}
\subsection{Scene Understanding module(SU) and Scale Transform module(ST)}\label{subsec3}
In this section, we will describe the SU and ST models in detail. \\
\textbf{Scene Understanding module (SU).} The SU model is proposed to deal with the loss of low-level feature problems. In the deeper network, the low-level features are always lost.  The monocular depth estimation is dense predicting vision tasks that need to use the deep CNNs to extract high-level features to build the mapping between RGB field and depth
field. So, the boundary blur is serious. To deal with this problem, we propose the SU to aggregate and learn the global scene information containing the low- and high-level features, and the global information will join in all encoding stages. The architecture of SU is shown in Figure 4. Firstly, we extract all scale feature maps from the backbone and then use two convolution layers with convolution kernel size=3 to compresses these feature maps to 64 channels. Secondly, we sample each feature map to the same resolution. Experiments finds that, if all feature maps sample to the second scale feature, the experiment result is best. Finally, we concatenate the feature maps and use a fusion layer to fuse these feature maps and compress it to 128 channels, the fusion layer includes two convolution layers with 5 and 3 kernel sizes separately. With the SU model, it aggregates all scale features from encoding stage, and then it learns the whole scene information and outputs the global information with 128 channels. This global information will join in every decoding stage.\\
\textbf{Scale Transform module (ST)}. In the decoder, the scale of each decoding phase is corresponding to the encoding phase with skip connection, so, before feeding the output of SU to decoder,  we should transform the global information to the suitable scale, thus, we design the ST to do this task. The architecture of ST is shown in Figure 5. In this module, we mainly use channel attention to set different weights to feature channels to adaptively change the feature to different scales. Firstly, we sample (or upsample) the feature map including global scene information outputted from SU to the corresponding scale in the decoder. Then, we compress the feature to 64 channels and use the average pooling layer to deal with the feature to a single pixel. Thirdly, we use the  kernel convolution layer to compress pooled feature to 32 channels and use the relu function to activate it. After that, we use the   convolution to recover the feature from 32 to 64 channels and use the sigmoid as the activate function. Finally, we produce a recovered feature with the feature which before pooling, and use   convolution to recover the feature to 128 channels. The processed feature maps will be transformed to the scale according to every phase of the decoder and be sent to the corresponding decoding step as a skip connection. The SU can learn the global information of the scene and the ST will transform the global information to each scale. The ST not only change the resolution of global information but also adaptively learn what features are needed in different decode scale and change the feature. With  ST to build pyramid reduces a lot of model params than pre-works, such as Chen et al. [13], Yang et al. [14]. The number of params of several models comparison results can be seen in the experiment section.
\begin{figure}%
\centering
\includegraphics[width=0.9\textwidth ]{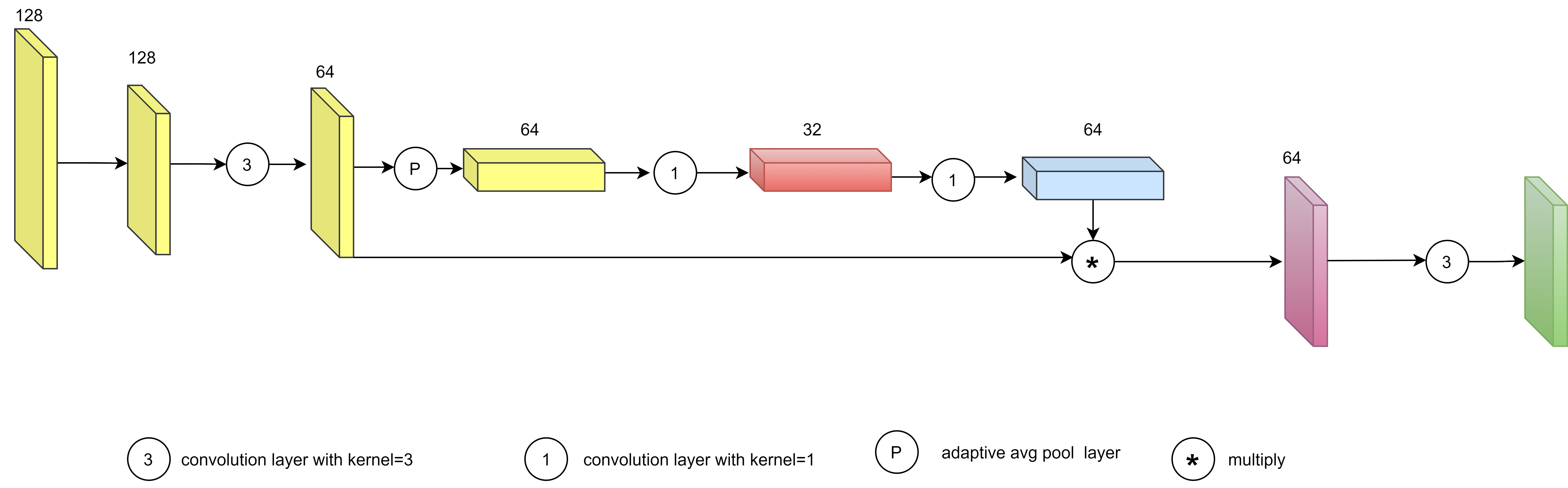}
\caption{The architecture of the Scale Transform module.}\label{fig5}
\end{figure}
\subsection{Boundary Aware Depth Loss(BAD)}\label{subsec3}
In this section, we propose a novel depth loss function Boundary Aware Depth loss (BAD) to pay attention to the usually ignored loss caused by the boundary in training.
Generally, the depth at the plane is continuous with a smoother gradient, while the depth at the boundary is discontinuous with a large gradient. The boundary area always occupies a little proportion though the gradient of boundary is larger, it causes little loss in training, so, it is easily ignored, especially in indoor scenes, there are more planes than ordinary scenes, due to the existence of walls, ceilings, tables, and beds etc. Therefore, the models tend to predict the depth of the entire scene with the same smooth depth when training the depth estimation model, which aggravates the boundary blur problem To deal with the ignored loss caused by the boundary, Hu et al. [10] proposed the gradients of depth item in total loss function to guide the model to predict the depth maps with accurate boundary gradient. However, the gradient loss term will not play an ideal role when the background and boundary depth is too large or too small at the same time. In this paper, base on Hu et al. [10], we propose a novel depth loss function BAD to pay attention to the boundary depth for improving the depth accuracy of object boundary. BAD guides the training process by setting a Boundary aware weight for each pixel. The BAD is defined as:
\begin{equation}
L_{BAD} =(1+\alpha\omega)(\ln(\lvert d-\hat{d} \rvert+0.5))  \label{eq1}
\end{equation}
  $L_{BAD}$  contains two items, the boundary aware weight and depth predict item.  Where the $\omega$ is boundary aware weight and  $d$ is the true depth,  $\hat{d}$ is predicted depth. The $\alpha$ is  aware factor and we set  $\alpha$ =0.3 and use the Sobel operator [22] to extract gradient in this paper. These pixels will be focused, if the boundary aware weight is big. Boundary aware weight  is define as:
  \begin{equation}
\omega =\frac{\ln(\lvert g_x\rvert+\lvert g_y\rvert+0.5)}{\frac{1}{N}\sum_{}(\lvert g_x\rvert+\lvert g_y\rvert)}
(\lvert g_x-\hat{g_x} \rvert+\lvert g_y-\hat{g_y} \rvert)  \label{eq2}
\end{equation}
  where the $g_x$ is gradient of x scale and the $g_y$ is gradient of y scale in the ground true. The  $\hat{g_x}$ is gradient of x scale and the $\hat{g_y}$  is the gradient of y scale in predicted depth maps. The $N$ is the total number of pixels. $\omega $ includes the true and error items,  will enforce the model to focus on boundary area by setting different weights to these pixels. The true item will become big when these pixels have a big gradient in the ground truth. The error items will show its role if there is a large gradient prediction error. When the background and boundary depth are too large or too small at the same time, that is the depth is false but the gradient is right, the gradient loss item [10] can’t aware this error, our true items and the depth error item in (1) will become big, which guides the model to focus these pixels together even though the gradient error is small. The error item is used to guide the model to focus these boundary field where the gradient error is big, the depth loss will be huge when the true item and the error item are big at the same time, that is our model predict the blurry depth in the boundary field. To ensure our model can take a bigger aware in the object boundary and point clouds quality, we retain the edge loss item and normal loss item proposed by [10]. The total loss is defined as:
  \begin{equation}
      L=L_{BAD}+L_{grad}+L_{normal} \label{eq3}
  \end{equation}
  Extensive experiments show that BAD can improve the accuracy of the boundary and depth obviously.
  \section{ Experiments}\label{sec4}
  In this section, we will introduce the evaluation indicators in our experiment firstly. Then we introduce the datasets used in our experiment, and finally, we conduct various experiments on the datasets to prove the effectiveness of the novel modules and functions.
\subsection{Quantitative evaluation indexes}\label{subsec4}
In this section, we will introduce the evaluation indicators in our experiment firstly. Then we introduce the datasets used in our experiment, and finally, we conduct various experiments on the datasets to prove the effectiveness of the novel modules and functions.\\
Root mean squared error (RMSE):\\
\begin{equation}
    RMSE=\sqrt{\frac{1}{\lvert T\rvert}\sum_{y\in T}\Vert\hat{g_i}-g_i\Vert^2} \label{eq4}
\end{equation}
Absolute relative difference (AbsRel):
\begin{equation}
    Abs Rel=\frac{1}{\lvert T\rvert }\sum_{y\in T}\lvert \hat{g_i}-g_i\rvert/g_i
\end{equation}
Log10:
\begin{equation}
    log10=\frac{1}{\lvert T\rvert}\sum_{y\in T}\lvert \log_{10}\hat{g_i}-\log_{10} g_i \rvert
\end{equation}
Threshold($\delta$):
\begin{equation}
    \% \, of \, y_i \, s.t \, max(\frac{\hat{g_i}}{g_i},\frac{g_i}{\hat{g_i}})=\delta<thr
\end{equation}
Where  $thr=1.25,1.25^2,1.25^3$ the  $g_i$is ground truth,  $\hat{g_i}$ predict depth value, and $T$ is the available pixels in the ground truth.
 \subsection{Datasets and experimental setting}\label{subsec4}
 This paper focuses on indoor scenes which have a complex scene structure with a large number of objects. So we mainly train and evaluate our model in NYU Depth v2 [23]. The NYU-Depth V2 [23] dataset is the most popular indoor dataset in monocular depth estimation and semantic segmentation. It uses a Kinect depth camera [24] to capture images, mainly for scene understanding. It contains 1449 pairs of RBGD image pairs with a resolution of 640 × 480 from 464 different indoor scenes from 3 cities. These image pairs are divided into two parts. 795 image pairs captured in 249 scenes are used as the training set, and 654 image pairs from 215 scenes are used as the test set. In addition, the data set also contains corresponding semantic segmentation label information. In our experiment, we also use the training dataset that contains 50K RGB-D images and was preprocessed by Hu et al. [10].In this paper, we use the Pytorch [25] to implement our model, and then in the encoder state, we use the SENet-154[21] as our backbone to initialize the pre-trained model by ImageNet [26]. We set the LR = 0.0001 and use the learning Adam optimizer. Furthermore, we set a rate decay policy to the learning rate by reducing it to 10 $\%$ every 5 epochs, and we set   $\beta_1$ = 0.9,$\beta_2$ = 0.999, epochs = 10, and weight decay as 10-5. Our model was trained and evaluated on a two-piece Tesla V 100(32GB version) with a patch size is 16.

\begin{table}[h]
\begin{center}
\begin{minipage}{270pt}
\caption{Evaluation results of depth estimation on the NYU V2 test set. The best results are boldfaced, and the second-best ones are underlined. The shown values of the evaluated methods are those reported by the authors in their paper.}\label{tab1}%
\begin{tabular}{ccccccc}
\toprule
Methods & $\delta_1 \uparrow$  & $\delta_2 \uparrow$ & $\delta_3 \uparrow$ & AbsRel $\downarrow$ & RMSE $\downarrow$ &log10 $\downarrow$\\
\midrule
Eigen et al. [3]& 0.611 & 0.887 & 0.971 & 0.215 & 0.907 & -\\
Liu et al. [27] & 0.614 & 0.883  & 0.971 &  0.230 & 0.824 & 0.095 \\
Cao et al. [28] & 0.646 & 0.892  & 0.968 & 0.232 &  0.819 & 0.063 \\
Li et al. [29]  & 0.788 & 0.958 & 0.991 & 0.143 & 0.635 & 0.063  \\
Laina et al. [30] & 0.811 & 0.953 & 0.988 & 0.127 & 0.573 & 0.055\\
Xu et al. [31] & 0.811 & 0.956 & 0.987 & 0.121 & 0.586 & 0.052 \\
Mal et al. [32] & 0.810 & 0959 & 0.989 & 0.143 & - & -\\
Lee et al. [33] & 0.815 & 0.963 & 0.991 & 0.139 & 0.572 & -\\
Hao et al. [34] & 0.828 & 0.965 & 0.992 & 0.127 & 0.555 & 0.053\\
Fu et al. [9]  & 0.828 & 0.965 & 0.992 & \underline{0.115} &\textbf{ 0.509} & 0.051\\
Qi et al. [35] & 0.834 & 0.960 & 0.990 & 0.128 & 0.569 & 0.057\\
Hu et al. [10] & 0.866 & 0.975 & 0.993 & \underline{0.115} & 0.530 & 0.050\\
chen et al. [13] &\textbf{ 0.878} & \textbf{0.977} & \textbf{0.994} &\textbf{ 0.111} & \underline{0.514} &\textbf{ 0.048} \\
Yang et al. [14] & 0.864 & 0.972 & 0.993 & \underline{0.115} & 0.525 & 0.050 \\
Our baseline & 0.856 & 0.972 & 0.992 & 0.120 & 0.541 & 0.056 \\
Ours & \underline{0.869} & \textbf{0.977} & \textbf{0.994} & \underline{0.115} & 0.519 & \underline{0.049} \\

\botrule
\end{tabular}

\end{minipage}
\end{center}
\end{table}
\noindent

\begin{figure}%
\centering
\includegraphics[width=0.9\textwidth ]{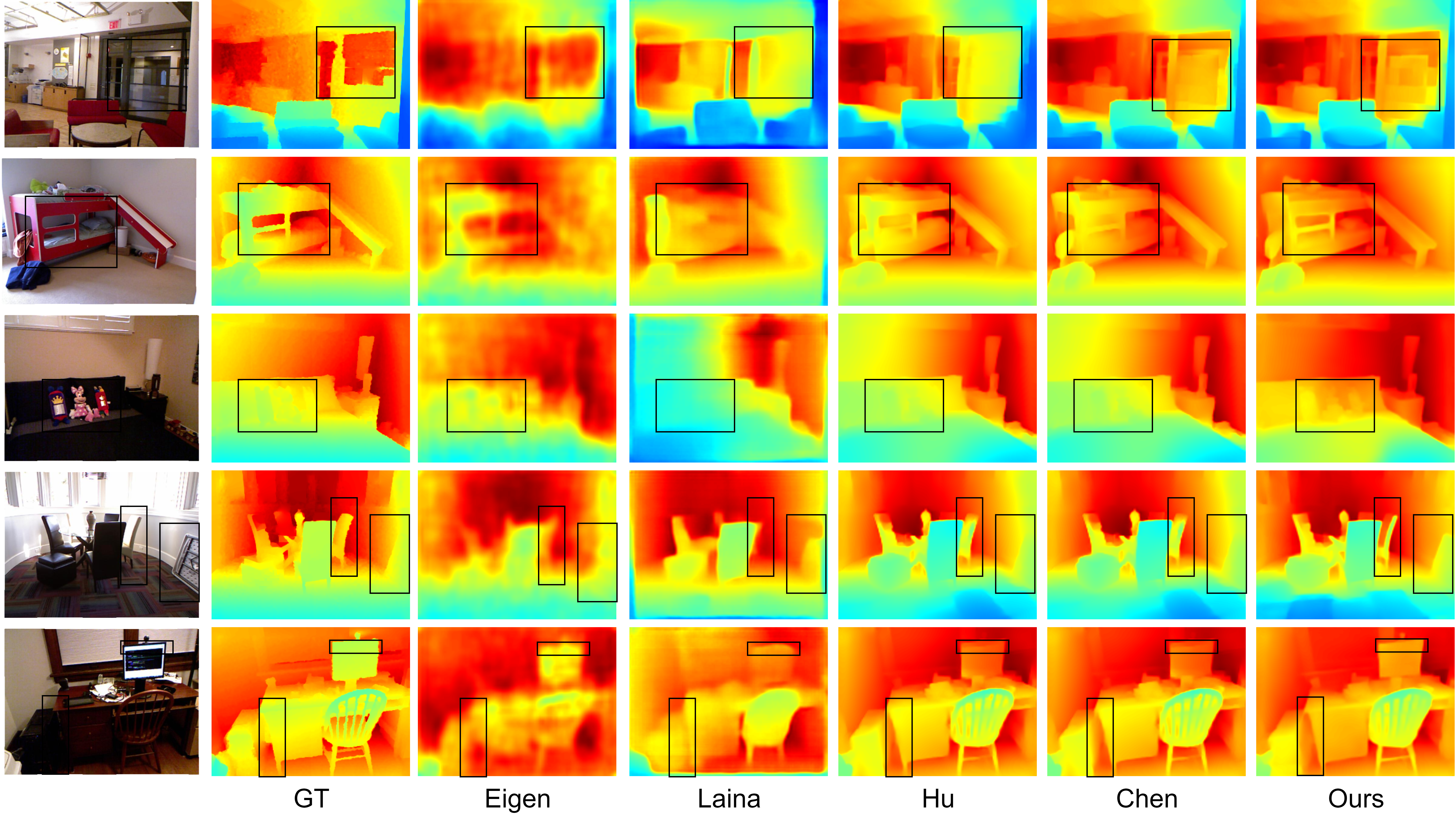}
\caption{Qualitative results on the NYU-D V2 test set. From left to right: input RGB images, ground-truth depth maps, results of Eigen et al. [3], Laina et al. [30], Hu et al. [10], Chen et al. [13], and our method respectively.}\label{fig6}
\end{figure}
\subsection{Performance comparisons}\label{subsec4}
In this section, we evaluate our model from both qualitative and quantitative points of view on the NYU-Depth V2 dataset[23]. Firstly, we compare the different state-of-the-art models with our model tested on the NYU-Depth V2 dataset[23] by the common indicators and the results are shown in From Table 1 we can see that our model shows state-of-the-art in  and  . Additionally, we also get the second state in  and AbsRel. Although our accuracy is next Chen et al. [13], [13] builds an FFP to get a fusion of features in each scale to improve the predicting accuracy, which leads to the model becoming big.  in contrast, we only fuse features once and use the ST to transform the fusion feature to different scales. What’s more, we only sample the feature 8 times (a SU 4 times and 4 ST 4 times ), but Chen et al. [13] sampled 25 times of the feature. Furthermore, the output of SU only has 128 channels, which also reduces the params Moreover, our main purpose of this paper is not to improve the accuracy of depth but to get clear boundaries in the depth maps. To prove our contribution in predicting more clear boundaries, we will provide the qualitative results of our model, which is shown in Figure 6, which includes five group results. From Figure 6, we can see our model predicts the more clear structure than other models,  even better than the ground truth, like the black boxes in the first group. Analysis finds that, the ground truth is captured by the Kinect [24] using an infrared ray, but the glass will reflect the ray which leads to the depth loss in the glass. But the monocular depth estimation can predict the depth about them relatively accurate. In the second group, our model predicts the best clear structure of bed with a sharp boundary. Although Chen et al. [13] and Hu et al. [10] keep the whole structure of the bed, they suffer from serious boundary smooth leading to some structures indistinguishable from the background. The same situation also appeared in the third group. And in third group,  Chen et al. [13] and Hu et al. [10] can’t predict the toy with a clear boundary that causes the toy and the sofa to be mixed, and the toy is completely invisible from the depth map. By contrast, our model predicts the toy obvious boundary, the toy and the sofa can be distinguished. In the fourth group, we can see that other methods predict fuzzy table boundaries, and with the boundary is sharp, especially in the area selected by the black box. Although Chen et al. [13] and Hu et al. [10] also predicted the overall structure of the chair, the predicted edges of the backrest were very blurry and there was a serious smoothing phenomenon. Additionally, in the black box on the wall, we can see a chair placed on the wall from ground truth. Our method not only successfully predicts chairs with sharp edges, but also suppresses the boundary smoothing phenomenon, so the chair and the wall are distinguished. But other algorithms did not deal with the boundary smoothing phenomenon, resulting in the wall and the chair being indistinguishable. The same situation also occurs in the fifth group, in which the other algorithms did not predict clear boundary between the cabinet and the table. Because the cabinet is back against the wall, the predicted depth of the display is very close to the depth of the wall. As a comparison, our model not only predicts clear boundaries, but also are more discriminatory than others in the depth prediction of the display. In general, our algorithm can preserve the overall structure of the scene better than other algorithms, and can effectively distinguish objects and background thanks to our depth maps with clear boundaries. In addition, clear boundary information also helps us to get more accurate depth predictions when the background and object depths are similar. 
\begin{figure}%
\centering
\includegraphics[width=0.9\textwidth ]{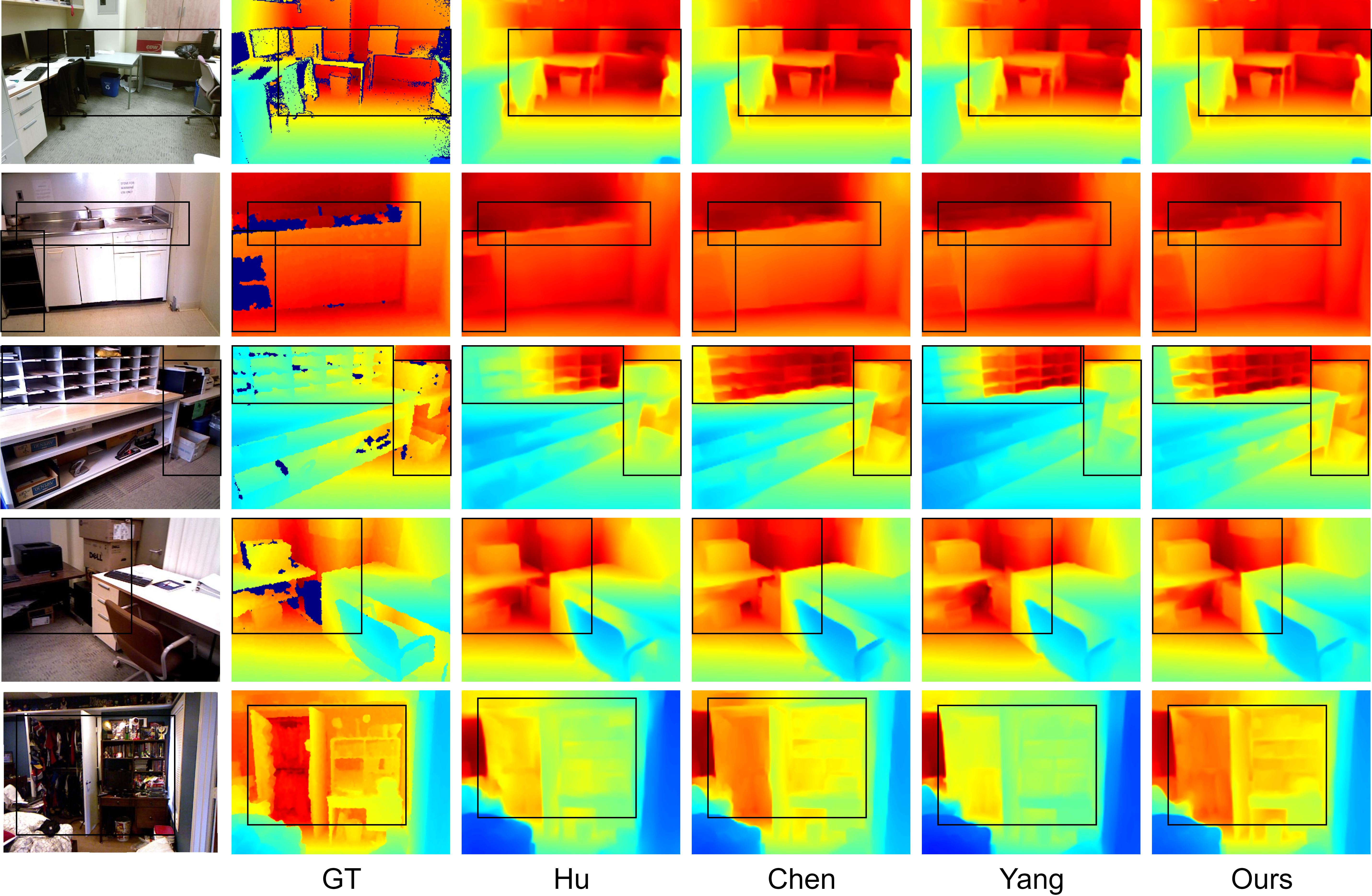}
\caption{Qualitative results of the test on the SUN RGB-D. From left to right: input RGB images, ground-truth depth maps, results of Hu et al. [10], Chen et al. [13], Yang et al. [14], and our method respectively.}\label{fig7}
\end{figure}
\subsection{Models test in other dataset}\label{subsec4}
To evaluate our model more thoroughly, we test our pre-trained model in the SUN RGB-D [36] directly. SUN RGB-D [36] is a scene understanding benchmark, which includes three datasets NYU-Depth V2 dataset[23], B3DO[37], SUN3D[38]. To ensure the effectiveness of the comparison, we choose Hu et al. [10], Chen et al. [13], Yang et al. [14] as the comparison models. We use the same dataset in the training process of the comparison models. l the dataset is processed by Hu et al. [10] base Depth V2 dataset[23]. The comparing results are shown in Figure 7. From Figure 7, we select five sets of depth maps predicted from different scenes as a comparison. In the first group, we can see the comparison models have preserved the object structure of the scene, but our model’s object boundaries are the clearest (tables and chairs in the black box). The lack of depth in ground truth is due to specular reflections of the scene, which makes it difficult to obtain the depth in the corresponding area. This is the disable in RGB-D camera and LIDAR. This phenomenon also can be found in the second and fourth groups. In the second group,  from the black boxes we can see that our model predicts a desktop with clear boundary. What’s more, we also predict the outline of the faucet but the other algorithms do not get a clear outline, and they are also difficult to distinguish the faucet from the background. From the third group, we can see that our algorithm retains the structure of the cabinet, while also obtaining sharp boundary in the left black box. Other algorithms retain the structure but without the clear boundary, and some cannot even retain the cabinet structure. ,and  we preserved the clearest structure of the cabinet again in the third group. In the fourth group, we can also see that a clearer object structure is preserved by ours than others. From the fifth group, we can see that our model shows obvious advantage over other algorithms in structure preservation, it retains the complete object structure while also getting sharp boundary. This makes the object can be perfectly distinguished from the background. In other algorithms,  it is difficult to distinguish the object from the background due to the blurry of the structure and smooth edges. Moreover, when the edges are blur, the object is easily mistaken as the background, and thus, it will has a depth value similar to the background, resulting in huge errors.
\subsection{Boundary accuracy comparisons}\label{subsec4}

\begin{table}
\begin{center}
\begin{minipage}{250pt}
\caption{Accuracy of recovered edge pixels in-depth maps under different thresholds. The best results are boldfaced, and the second-best ones are underlined.}\label{tab2}%
\begin{tabular}{@{}ccccc@{}}
\toprule
Thres & Method  & Prec $\uparrow$  & Recall $\uparrow$ & F1 $\uparrow$ \\
\midrule
\multirow {7}{*}{\centering 0.25}    & Laina et al. [30] & 0.489 & 0.435 & 0.454  \\
    & Xu et al. [17] & 0.516 & 0.400 & 0.436  \\
    & Fu et al. [9] & 0.320  & \textbf{0.583} & 0.402  \\
    & Hu et al. [10] & 0.644 & 0.508 & 0.562 \\
    & Chen te al. [13] & 0.645 & 0.520 &\underline{0.570} \\
    & Yang et al. [14] &\textbf{ 0.652} & 0.518 &\underline{0.570} \\
    & Our & \underline{0.651} & \underline{0.524} &\textbf{ 0.574}\\
\multirow {7}{*}{\centering 0.5}    & Laina et al. [30] & 0.536 & 0.422 & 0.463  \\
    & Xu et al. [17] & 0.600 & 0.366 & 0.439  \\
    & Fu et al. [9] & 0.316  &0.473 & 0.412  \\
    & Hu et al. [10] & 0.668 & 0.505 & 0.568 \\
    & Chen te al. [13] & 0.663 & \textbf{0.523} & \underline{0.578} \\
    & Yang et al. [14] &\textbf{ 0.685} & 0.510 & 0.576 \\
    & Our & \underline{0.680} & \underline{0.520} &\textbf{ 0.582}\\    
\multirow {7}{*}{\centering 1}    & Laina et al. [30] & 0.670 & 0.479 & 0.548 \\
    & Xu et al. [17] & \textbf{0.794} & 0.407 & 0.525  \\
    & Fu et al. [9] & 0.483  &0.512 & 0.485  \\
    & Hu et al. [10] & 0.759 & 0.540 &0.623 \\
    & Chen te al. [13] & 0.749 & \textbf{0.554} & 0.630 \\
    & Yang et al. [14] & \underline{0.774} & 0.544 & \underline{0.631} \\
    & Our & 0.770 & \underline{0.553} & \textbf{0.635}\\    
    
\botrule
\end{tabular}

\end{minipage}
\end{center}
\end{table}

\noindent
In order to more effectively prove that our model can predict more accurate boundaries than others, we compare them on specific data. We follow Hu et al. [10] using the Precision, Recall, and F1 scores to evaluate the method performance, the results show in Table 2, from which it can be found that our algorithm has achieved 3 SOTA and 5 sub-SOTA in 3 indicators with 3 different thresholds. The results proves that our model performs SOTA in edge accuracy.
\subsection{Generating point cloud from depth map}\label{subsec4}

\begin{figure}[h]%
\centering
\includegraphics[width=1\textwidth ]{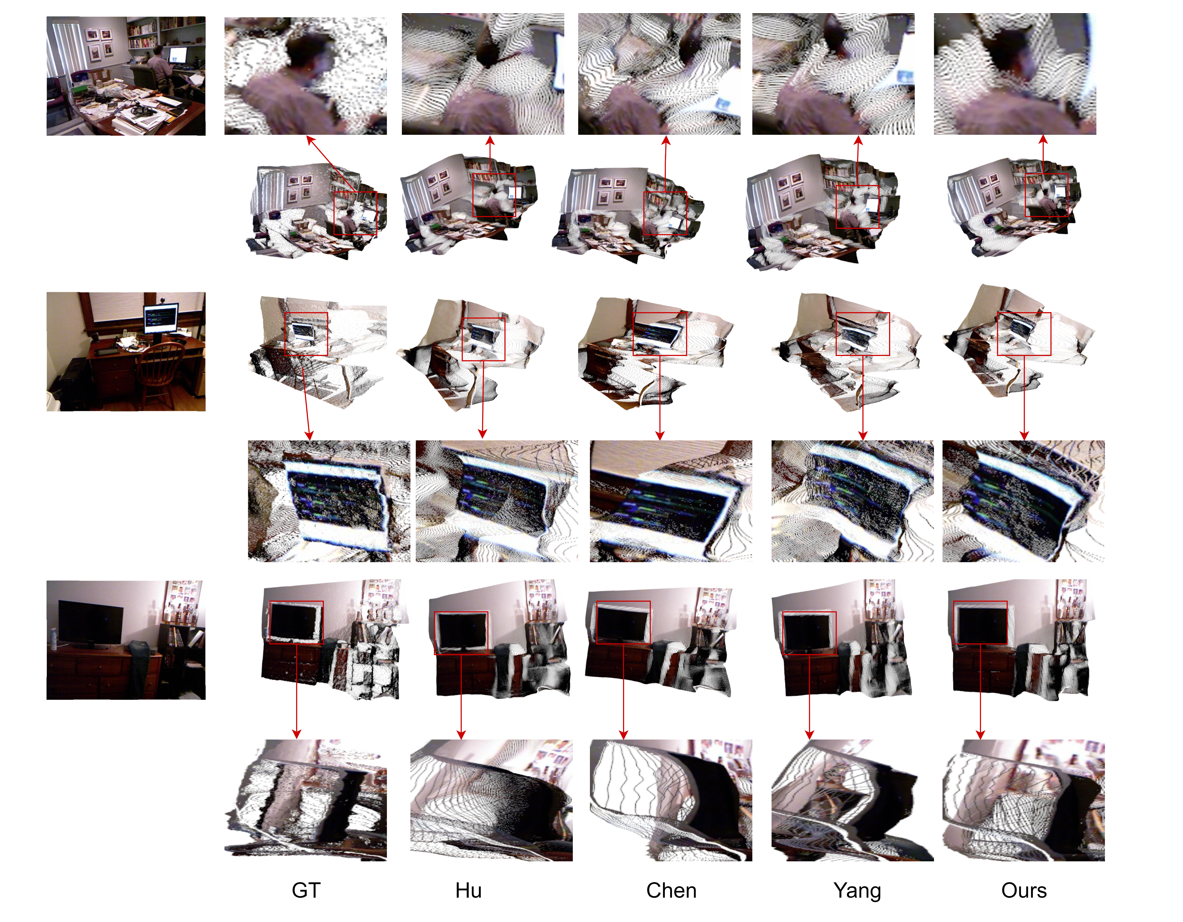}
\caption{Qualitative results of the test on the SUN RGB-D. From left to right: input RGB images, ground-truth depth maps, results of Hu et al. [10], Chen et al. [13], Yang et al. [14], and our method respectively.}\label{fig8}
\end{figure}
As mentioned in the pre-chapters, the Sharpe boundary can effectively suppress " flying pixels "in the point clouds projected by depth maps. To prove this conclusion, we compare the point clouds projected by our method with others. The results are shown in Figure 8. From the first group, other algorithms have serious pixel drift in the head of the character in the red block. Our algorithm suppresses this phenomenon well. The same situation also appeared in the second set. The projection effect of Hu et al. [10] is relatively good, but it still appears seriously at the upper boundary of the screen. Compared with others, our model also has distortions at the bottom edge of the screen, but the overall structure is better than others. In the third group, we can see that each method has better preserved the overall structure of the scene. However, by changing the perspective, we found that the TV screen predicted by other methods has serious" flying pixels " and the screen is distorted which surface is curving. Although ours suffer a slight of flying pixels at the top of the screen, the overall screen is not distorted. Through point clouds comparison experiments, we proved that our algorithm can effectively suppress the phenomenon of "flying pixels", and also proved the perspective that the accurate edge information can help to improve the quality of the point clouds predicted from depth maps. 
\begin{figure}[h]%
\centering
\includegraphics[width=1\textwidth ]{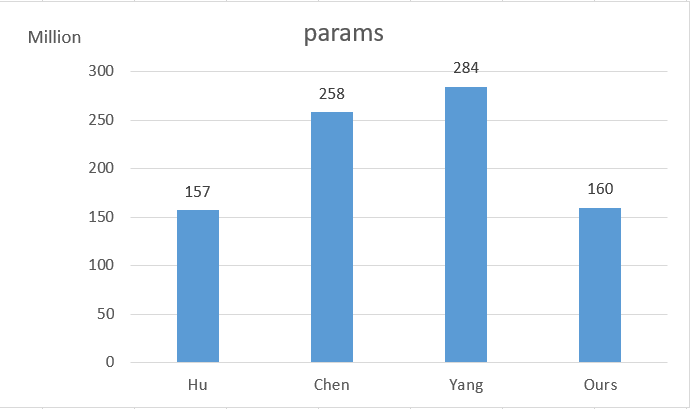}
\caption{Compare the number of model’s params. From left to right: results of Hu et al. [10], Chen et al. [13], Yang et al. [14], and our method respectively.}\label{fig9}
\end{figure}
\subsection{The comparison of model’s params}\label{subsec4}
In order to prove the ST can show a great performance in reducing the number of params, we make the comparison between ours and others and the comparison results can be seen in FIGURE 9. From Figure 9, we find that our model’s params are slightly more than Hu et al. [10],  that is because [10] just fused the multi-scale feature one time on a single scale and did not change the fusion feature to different scales. But our model not only fuses the all-scale feature in every training process but also transforms the fusion feature into 5 different scales. What’s more, Comparing with Chen et al. [13] and Yang et al. [14] which build the FFP, our model only contains two-thirds of params of these FFP models.
\begin{table}[h]

\begin{minipage}{400pt}
\caption{Depth accuracy comparison with our module or not. The best results are boldfaced. }\label{tab3}%
\begin{tabular}{@{}ccccccc@{}}
\toprule
Methods & $\delta_1 \uparrow$  & $\delta_2 \uparrow$ & $\delta_3 \uparrow$ & AbsRel $\downarrow$ & RMSE $\downarrow$ &log10 $\downarrow$  \\
\midrule
Baseline & 0.856   & 0.972  & 0.993 &  0.120 & 0.541 & 0.051\\
Baseline+ BAD &\textbf{ 0.870} & 0.974 & 0.993 & \textbf{0.115} & 0.525 & \textbf{0.049}   \\
Baseline+ BAD+SU+upsamle directly & 0.863 & 0.975 & 0.993 & 0.117 & 0.523 & 0.050      \\
Baseline+ SU+ST  & 0.863 & 0.975 & 0.993 & 0.119 & 0.524 & 0.050 \\
Baseline+ SU+ST+BAD & 0.869 & \textbf{0.977} & \textbf{0.994} & \textbf{0.115 }& \textbf{0.519 }&\textbf{ 0.049}  \\

\botrule
\end{tabular}
\end{minipage}

\end{table}

\noindent

\begin{table}[h]
\begin{center}

\begin{minipage}{250pt}
\caption{Boundary accuracy comparison with our module or not. The best results are boldfaced. ( thresholds=0.5)}\label{tab4}%
\begin{tabular}{@{}cccc@{}}
\toprule
Methods & Prec$\uparrow$  & Recall$\uparrow$ & F1$\uparrow$\\
\midrule
Baseline  & 0.678 & 0.503 & 0.569 \\
Baseline+ BAD  & 0.672 & 0.508 & 0.571   \\
Baseline+ BAD+SU+upsample directly & 0.676 & \textbf{0.520} & 0.580     \\
Baseline+ SU+ST   & 0.671 & 0.515 & 0.575\\
Baseline+ SU+ST+BAD  &\textbf{ 0.680} &\textbf{ 0.520} & \textbf{0.582} \\

\botrule
\end{tabular}
\end{minipage}
\end{center}
\end{table}

\noindent
\subsection{Ablation studies}\label{subsec4}
To explore our model in detail, we designed corresponding 
ablation experiments for the proposed method. The results are shown in Table 3 and Table 4 base on thresholds=0.5. The first set of experiments is mainly to show the performance of the baseline, and use the estimated result as the benchmark for subsequent comparisons. For the baseline, we use SeNet154 [21] as the backbone and use the composite loss function proposed by Hu et al. [10] to supervise the model training. From the results of the second group, it can be known that directly adding BAD to supervision model training can improve the prediction accuracy of the model effectively. To prove the effectiveness of ST, the third group adds SU base the second group, and, directly sample the output of SU without ST., Comparing the third group and the fifth group, we can see that direct upsampling global scene information will impact the enhance from SU. At the same time, by comparing with the second group, it can be concluded that SU has a significant improvement in the edge accuracy of the model. Comparing the baseline and the fourth group, we can prove the effectiveness of the SU+ST model. Comparing the baseline with the fifth group proves the effectiveness of BAD.

\section{Conclusion}\label{sec5}
 To deal with the boundary blurry caused by low-level information loss in process of feature extracting and  boundary smooth in the boundary area during training process, a Scene Understanding module, and Scale Transform module and a Boundary Aware Depth loss function were proposed. In the Scene understanding module and Scale Transform module, we focus on  dealing with the information loss. The Scene understanding module can learn the global information about the scene and the Scale Transform module will transform the global scene information to multi-scale for building a feature pyramid with little additional params. The Boundary Aware Depth loss was design to enforce the model pay attention to the depth in the boundary field in the model training. The extensive experiments show our modules and the novel loss functions ensure our model can predict depth maps with more clear object boundaries than others. The most important is the object cannot be predicted with the depth value same as the background without a clear boundary, that is to say, the boundary information can influent the depth prediction. But some problems still exist, for example, although our model can recover the boundary very well, the point clouds are not always nice enough,  like some planes are not smooth.  the other problem is the complex time about our model is high.. In future work, we will focus on how to improve the accuracy of depth prediction and reduce the time complexity. Also, we will further explore the influence of object boundaries on depth prediction.
\section{References}\label{sec6}
[1]	F. Guo, J. Tang, and H. Peng, "Adaptive estimation of depth map for two-dimensional to three-dimensional stereoscopic conversion," Optical Review, vol. 21, no. 1, pp. 60-73, 2014.\newline
[2]	C. Tang, C. Hou, and Z. Song, "Depth recovery and refinement from a single image using defocus cues," Journal of Modern Optics, vol. 62, no. 6, pp. 441-448, 2015.\newline
[3]	D. Eigen, C. Puhrsch, and R. Fergus, "Depth map prediction from a single image using a multi-scale deep network," arXiv preprint arXiv:1406.2283, 2014.\newline
[4]	D. Eigen and R. Fergus, "Predicting depth, surface normals and semantic labels with a common multi-scale convolutional architecture," in Proceedings of the IEEE international conference on computer vision, 2015, pp. 2650-2658.\newline
[5]	F. Liu, C. Shen, G. Lin, and I. Reid, "Learning depth from single monocular images using deep convolutional neural fields," IEEE transactions on pattern analysis and machine intelligence, vol. 38, no. 10, pp. 2024-2039, 2015.\newline
[6]	A. Sagar, "Monocular depth estimation using multi scale neural network and feature fusion," arXiv preprint arXiv:2009.09934, 2020.\newline
[7]	K. Wu, S. Zhang, and Z. Xie, "Monocular depth prediction with residual DenseASPP network," IEEE Access, vol. 8, pp. 129899-129910, 2020.\newline
[8]	J. Jiao, Y. Cao, Y. Song, and R. Lau, "Look deeper into depth: Monocular depth estimation with semantic booster and attention-driven loss," in Proceedings of the European conference on computer vision (ECCV), 2018, pp. 53-69.\newline
[9] H. Fu, M. Gong, C. Wang, K. Batmanghelich, and D. Tao, "Deep ordinal regression network for monocular depth estimation," in Proceedings of the IEEE conference on computer vision and pattern recognition, 2018, pp. 2002-2011.\newline
[10] J. Hu, M. Ozay, Y. Zhang, and T. Okatani, "Revisiting single image depth estimation: Toward higher resolution maps with accurate object boundaries," in 2019 IEEE Winter Conference on Applications of Computer Vision (WACV), 2019: IEEE, pp. 1043-1051.\newline
[11]	K. Swami, P. V. Bondada, and P. K. Bajpai, "ACED: Accurate And Edge-Consistent Monocular Depth Estimation," in 2020 IEEE International Conference on Image Processing (ICIP), 2020: IEEE, pp. 1376-1380.\newline
[12]	M. Reynolds, J. Doboš, L. Peel, T. Weyrich, and G. J. Brostow, "Capturing time-of-flight data with confidence," in CVPR 2011, 2011: IEEE, pp. 945-952.\newline
[13]	X. Chen, X. Chen, and Z.-J. Zha, "Structure-aware residual pyramid network for monocular depth estimation," arXiv preprint arXiv:1907.06023, 2019.\newline
[14]	X. Yang, Q. Chang, X. Liu, S. He, and Y. Cui, "Monocular Depth Estimation Based on Multi-Scale Depth Map Fusion," IEEE Access, vol. 9, pp. 67696-67705, 2021.\newline
[15]	B. Li, C. Shen, Y. Dai, A. Van Den Hengel, and M. He, "Depth and surface normal estimation from monocular images using regression on deep features and hierarchical crfs," in Proceedings of the IEEE conference on computer vision and pattern recognition, 2015, pp. 1119-1127.\newline
[16]	E. Ricci, W. Ouyang, X. Wang, and N. Sebe, "Monocular depth estimation using multi-scale continuous crfs as sequential deep networks," IEEE transactions on pattern analysis and machine intelligence, vol. 41, no. 6, pp. 1426-1440, 2018.\newline
[17]	D. Xu, W. Wang, H. Tang, H. Liu, N. Sebe, and E. Ricci, "Structured attention guided convolutional neural fields for monocular depth estimation," in Proceedings of the IEEE conference on computer vision and pattern recognition, 2018, pp. 3917-3925.\newline
[18]	M. Heo, J. Lee, K.-R. Kim, H.-U. Kim, and C.-S. Kim, "Monocular depth estimation using whole strip masking and reliability-based refinement," in Proceedings of the European Conference on Computer Vision (ECCV), 2018, pp. 36-51.\newline
[19]	S. F. Bhat, I. Alhashim, and P. Wonka, "Adabins: Depth estimation using adaptive bins," in Proceedings of the IEEE/CVF Conference on Computer Vision and Pattern Recognition, 2021, pp. 4009-4018.\newline
[20]	S. Liu, D. Huang, and Y. Wang, "Learning spatial fusion for single-shot object detection," arXiv preprint arXiv:1911.09516, 2019.\newline
[21]	J. Hu, L. Shen, and G. Sun, "Squeeze-and-excitation networks," in Proceedings of the IEEE conference on computer vision and pattern recognition, 2018, pp. 7132-7141.\newline
[22]	N. Kanopoulos, N. Vasanthavada, and R. L. Baker, "Design of an image edge detection filter using the Sobel operator," IEEE Journal of solid-state circuits, vol. 23, no. 2, pp. 358-367, 1988.\newline
[23]	N. Silberman, D. Hoiem, P. Kohli, and R. Fergus, "Indoor segmentation and support inference from rgbd images," in European conference on computer vision, 2012: Springer, pp. 746-760.\newline
[24]	Z. Zhang, "Microsoft kinect sensor and its effect," IEEE multimedia, vol. 19, no. 2, pp. 4-10, 2012.\newline
[25]	A. Paszke et al., "Pytorch: An imperative style, high-performance deep learning library," Advances in neural information processing systems, vol. 32, pp. 8026-8037, 2019.\newline
[26]	J. Deng, W. Dong, R. Socher, L.-J. Li, K. Li, and L. Fei-Fei, "Imagenet: A large-scale hierarchical image database," in 2009 IEEE conference on computer vision and pattern recognition, 2009: Ieee, pp. 248-255.\newline
[27]	F. Liu, C. Shen, and G. Lin, "Deep convolutional neural fields for depth estimation from a single image," in Proceedings of the IEEE conference on computer vision and pattern recognition, 2015, pp. 5162-5170.\newline
[28]	Y. Cao, Z. Wu, and C. Shen, "Estimating depth from monocular images as classification using deep fully convolutional residual networks," IEEE Transactions on Circuits and Systems for Video Technology, vol. 28, no. 11, pp. 3174-3182, 2017.\newline
[29]	J. Li, R. Klein, and A. Yao, "A two-streamed network for estimating fine-scaled depth maps from single rgb images," in Proceedings of the IEEE International Conference on Computer Vision, 2017, pp. 3372-3380.\newline
[30]	I. Laina, C. Rupprecht, V. Belagiannis, F. Tombari, and N. Navab, "Deeper depth prediction with fully convolutional residual networks," in 2016 Fourth international conference on 3D vision (3DV), 2016: IEEE, pp. 239-248.\newline
[31]	D. Xu, E. Ricci, W. Ouyang, X. Wang, and N. Sebe, "Multi-scale continuous crfs as sequential deep networks for monocular depth estimation," in Proceedings of the IEEE conference on computer vision and pattern recognition, 2017, pp. 5354-5362.\newline
[32]	F. Ma and S. Karaman, "Sparse-to-dense: Depth prediction from sparse depth samples and a single image," in 2018 IEEE international conference on robotics and automation (ICRA), 2018: IEEE, pp. 4796-4803.\newline
[33]	J.-H. Lee, M. Heo, K.-R. Kim, and C.-S. Kim, "Single-image depth estimation based on fourier domain analysis," in Proceedings of the IEEE Conference on Computer Vision and Pattern Recognition, 2018, pp. 330-339.\newline
[34]	Z. Hao, Y. Li, S. You, and F. Lu, "Detail preserving depth estimation from a single image using attention guided networks," in 2018 International Conference on 3D Vision (3DV), 2018: IEEE, pp. 304-313.\newline
[35]	X. Qi, R. Liao, Z. Liu, R. Urtasun, and J. Jia, "Geonet: Geometric neural network for joint depth and surface normal estimation," in Proceedings of the IEEE Conference on Computer Vision and Pattern Recognition, 2018, pp. 283-291.\newline
[36]	S. Song, S. P. Lichtenberg, and J. Xiao, "Sun rgb-d: A rgb-d scene understanding benchmark suite," in Proceedings of the IEEE conference on computer vision and pattern recognition, 2015, pp. 567-576.\newline
[37]	A. Janoch et al., "A category-level 3d object dataset: Putting the kinect to work," in Consumer depth cameras for computer vision: Springer, 2013, pp. 141-165.\newline
[38]	J. Xiao, A. Owens, and A. Torralba, "Sun3d: A database of big spaces reconstructed using sfm and object labels," in Proceedings of the IEEE international conference on computer vision, 2013, pp. 1625-1632.\newline




\end{document}